\documentclass[11pt]{article}

\usepackage{amsmath}
\usepackage{amsfonts}
\usepackage{amssymb}
\usepackage{mathrsfs,amsmath}
\usepackage{graphicx} 
\usepackage{epstopdf}
\usepackage{tabulary}
\usepackage{color}
\usepackage[table]{xcolor}
\usepackage{natbib}
\usepackage{url}

\begin{document}

\author{
  Reza Shahbazi\footnote{CK-12 data science team} \footnote{A Python implementation of this algorithm is available on GitHub at...}  (reza@ck12.org),
  Miral Shah (miral@ck12.org),
  Neeru Khosla (neeru@ck12.org)\\
  Palo Alto, CA, 94303, USA 
}

\title{Invariant Representation of Mathematical Expressions}

\maketitle

\begin{abstract}

While there exist many methods in machine learning for comparison of letter string data, most are better equipped to handle strings that represent natural language, and their performance will not hold up when presented with strings that correspond to mathematical expressions. Based on the graphical representation of the expression tree, here we propose a simple method for encoding such expressions that is only sensitive to their structural properties, and invariant to the specifics which can vary between two seemingly different, but semantically similar mathematical expressions.
  
\end{abstract}

\section{Introduction}
In comparison of educational texts, one will often have to make measurements of similarity between documents that contain, in addition to natural language, mathematical expressions. For example, we might be interested in matching the statement "Students must understand the properties of algebraic expressions of the form $(x+y)^2$" to a set of documents each with instructions on a different kind of algebraic expression. This can be problematic because NLP algorithms are generally better suited to process natural language and may not readily accept math on their input. To fix this, we can attempt to modify our approach to accommodate the mathematical inputs. For instance, we may assign a unique token to each mathematical expression, and replace it in document with the assigned token. The result will be a document consisting only of tokens, and can be processed with any number of NLP algorithms.

However, the above strategy will not be quite effective because unlike natural language where the tokens are invariantly represented (e.g. every instance of the word ``horse'' will be recognized as the same), mathematical expressions are unlikely to have an invariant representation. For instance, the expression $(x+y)^2$ may appear as $(\alpha + \beta)^2$ in some documents and $a^2 + b^2 + 2ab$ in others. For our measurements of similarity to be effective, ideally we would like to be able to recognize equivalent mathematical expressions as same  in all the various guises they might take on. This paper details our proposed algorithm for building invariant representations for mathematical expressions. In particular we convert mathematical expressions into their equivalent graphical representation, where the question of their identity is handled through graph isomorphism.

Our approach bears resemblances to graph canonization method \citep{arvind2008logspace} used in, for instance, molecule classification \citep{weininger1989smiles} where the features of interest are in the topology of the raw representations.

Considering the prevalence of neural network approaches, one might wonder how such an approach would fare in the current context. Indeed, recurrent neural networks and attention based architectures have proven quite useful in encoding sequential syntactic data, most notably, natural language \citep{devlin2018bert}. We feel that the statistical nature of the inferences made by neural networks does not reflect well the crystalline nature of mathematical deduction. While the two expressions in \ref{eq:ab_vs_xy} are absolutely the same, a statistical inference engine would assign a probability distribution that spans these, but also many other relevant, but not identical, expressions, which brings us to the next point. Our goal here is to build invariant representations, not to build a network of similarities between contexually or syntactically relevant expressions. The latter can certainly benefit from a neural network approach. The former seems to us better handled with a deterministic approach.

\section{Motivation}
Looking at the following two expressions, we can tell that they are expressing the same symbolic mathematical relationship
\footnote{Of course if we were to specify them further, they may become different expressions. For instance, if one were defined only on complex numbers and the other on positive integers, then obviously they wouldn't be the same.}, even though they look quite different.

\begin{equation}
\begin{split}
& a^2 + b^2 + 2ab \\
& (x + y)^2
\end{split}
\label{eq:ab_vs_xy}
\end{equation}

How do we capture this similarity? Part of the problem is that the same expression can appear in factored form, $(a+b)^2$, or expanded form, $a^2 + b^2 + 2ab$. Thankfully, there exist libraries and platforms that can easily rewrite the expression from one form to the other (e.g. Maple and SymPy). Therefore, as long as we consistently encode every expression in the same form, this should not be of concern. The more cumbersome issue is that one expression is stated in terms of $a$ and $b$ while the other is expressed via $x$ and $y$. Since we already have the means of resolving the identity, let us reformulate the problem and focus on the issue that requires our attention:

\begin{equation}
\begin{split}
& a^2 + b \\
& x + y^2
\end{split}
\label{xy_vs_ab}
\end{equation}

It might be tempting to tackle the issue by deciding on a set of canonical variables, $\{v_1,v_2..\}$ and rewrite every expression in terms of them. However, this method is bound to fail, for instance, when faced with $v_1^2 + v_2$ versus $v_2^2+v_1$.

We note that the isomorphy of the two expressions is in the structure of the algebraic operation they suggest, which in words would be something along "one variable plus the square of another variable". Therefore, what we need is to encode the expression in such a way that exposes this structure.

\begin{figure}[htp]
\includegraphics[trim=50 50 50 50, clip, scale=.5]{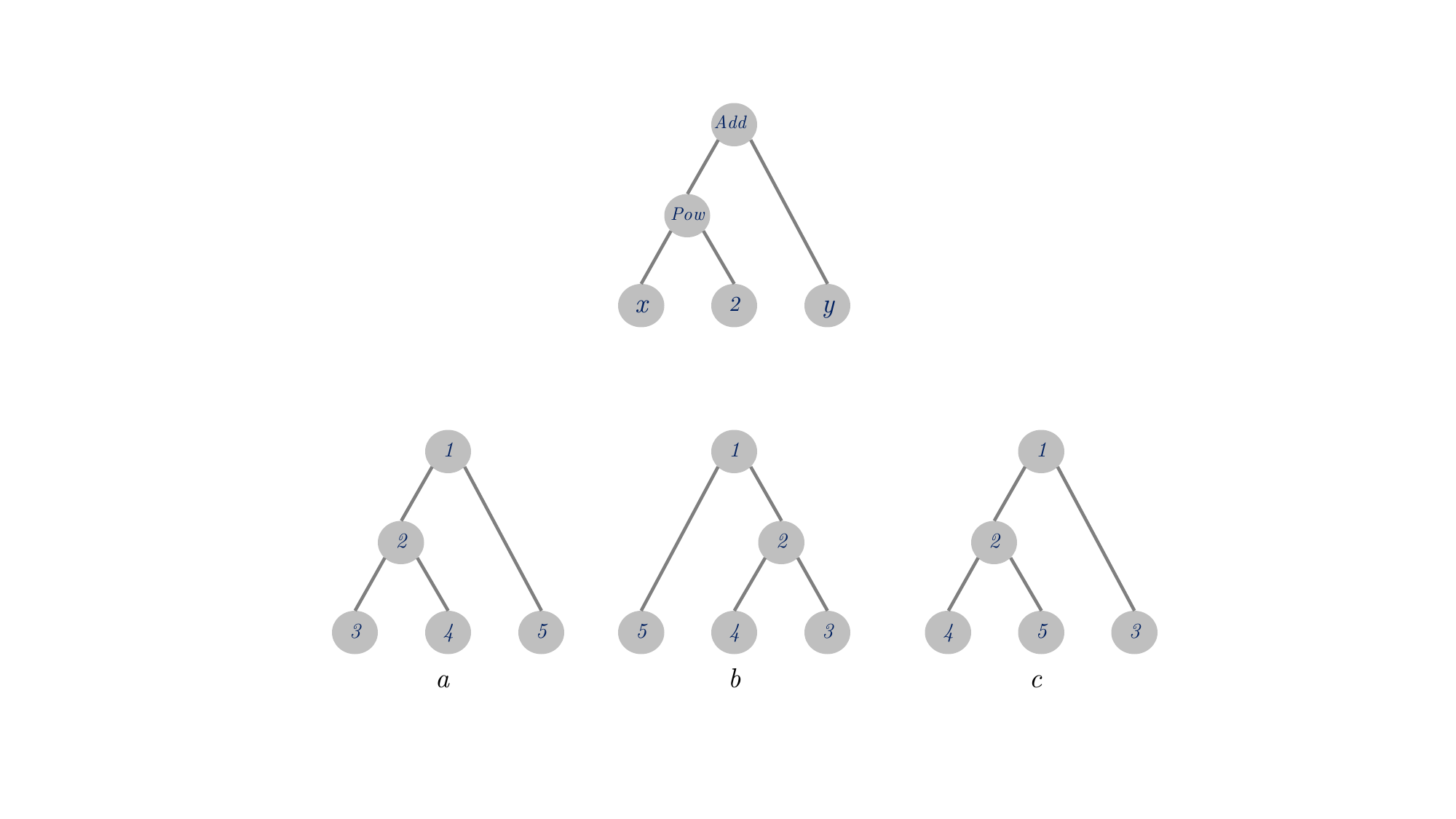}
\caption{Top: The expression tree for $x^2 + y$. Bottom: a and b are isomorphic, even though they appear different. Neither is isomorphic with c.}
\label{fig:isographs}
\end{figure}

We propose to use a modified form of the expression tree for this purpose, namely, a convenient way to represent the structure of a mathematical expression. The useful property of such a graph is that it makes explicit the property of interest, which is otherwise only implied in the nominal representation of the above expressions. Once we have the desired graph, then the problem of matching two expressions changes to that of graph isomorphism. Graph isomorphism is the well established problem of deciding whether two different graphs employ the same toplogical structure, regardless of their apparently different representation. For instance, in figure \ref{fig:isographs}, bottom, a and b are isomorphic to each other, but neither is isomorphic to c.

It should be noted that the question of whether two mathematical expressions are the same amounts to asserting the truth value of a formal statement, which is in general undecidable. Therefore, the algorithm proposed here would be better considered as a heuristic which can handle most ordinary cases likely to appear in school textbooks, but will nevertheless fail in the face of self referential statements. To make matters worse, in the proposed approach, once the expressions are converted into their graphical representation, their equality boils down to isomorphism of their corresponding graphs, which is conjectured to be NP-complete. Therefore, this approach may not be suitable for long and complex expressions that result in large graphs with many edges. In our experiments, the only times we observed the consequences of the computational complexity of isomorphism algorithms was in dealing with large and complex proof statements.

\section{Algorithm}
We start by parsing the expression and building an expression tree. Expression trees are undirected acyclic graphs built from a prefix notation of the expression. Specifically, we are interested in the algebraic expression trees. Figure \ref{fig:isographs}, top, illustrates an example of an expression tree for $x^2 + y$. To facilitate our goal of recognizing identical symbolic expressions, this initial tree is modified from its original form in the following ways: 

\begin{itemize}

\item While it is common for these trees to be binary, here we are relaxing this constraint and allowing each sub-tree to have more than two children. However, note that since $T(c_0,c_1,...c_n)$, an N-ary branch, can be rewritten as $T_0(T_1(...T_{n-1}(c_{n-1},c_n),c_1),c_0)$, a binary sub-tree, our discussion here extends to binary trees too. 

\item In a strict tree every node, including the leaves, can have at most one parent. In our graph we let leaves have multiple parents. The purpose of this is to ensure that all operations involving the same symbol terminate on it, as discussed momentarily.

\end{itemize}

\begin{figure}[htp]
\includegraphics[trim= 50 50 50 50, clip ,scale=.5]{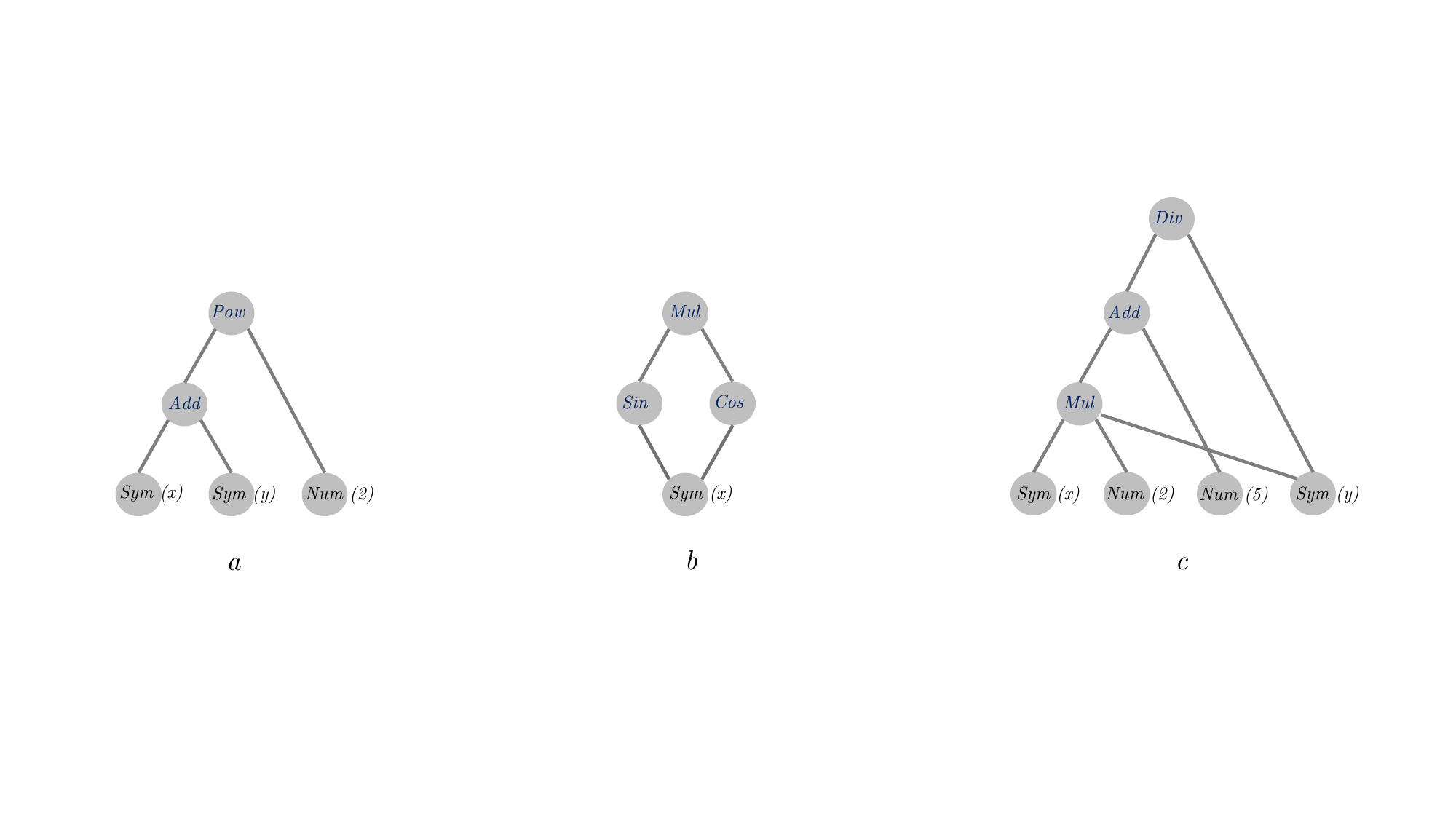}
\caption{Examples of expression trees. Parenthetic values next to the leaves are for visualization only, and are not part of the tree. \textit{a}: $(x+y)^2$; \textit{b}: $Sin(x)Cos(x)$; \textit{c}: $\frac{2xy+5}{y}$}
\label{fig:example_expressions}
\end{figure}

With the expression tree built, the adjacency matrix\footnote{A common representation for graphs is a square matrix with as many rows and columns as the number of vertices in the graph. If there is an edge connecting vertex $i$ to vertex $j$ then the $ij^{th}$ entry in the matrix will be $1$, otherwise $0$. This is called the adjacency matrix.}
together with the set of vertices will uniquely identify a symbolic expression. 
In our tree representation, the leaves are labeled ``Sym'' when they represent a symbol, and ``Num'' when they represent a number. Note that there is no need to differentiate multiple Sym leaves (for instance by subscripting them: $\text{Sym}_{1},\text{Sym}_{2},...$) because all the operations performed on the same symbol, terminate on the same Sym leaf. In other words two different leaves labeled Sym correspond to two different symbols (e.g. $x$ and $y$). See figure \ref{fig:example_expressions} for examples.

From there, the proper way to test whether two expressions thus coded are equal, will be to test whether they have the same vertices, and more importantly, if their graphs are isomorphic. Most graph libraries come equipped with isomorphism test functionality and the reader can pick their favorite library. For our experiments we use the NetworkX library for Python. 


\begin{figure}[htp]
\includegraphics[scale=.5]{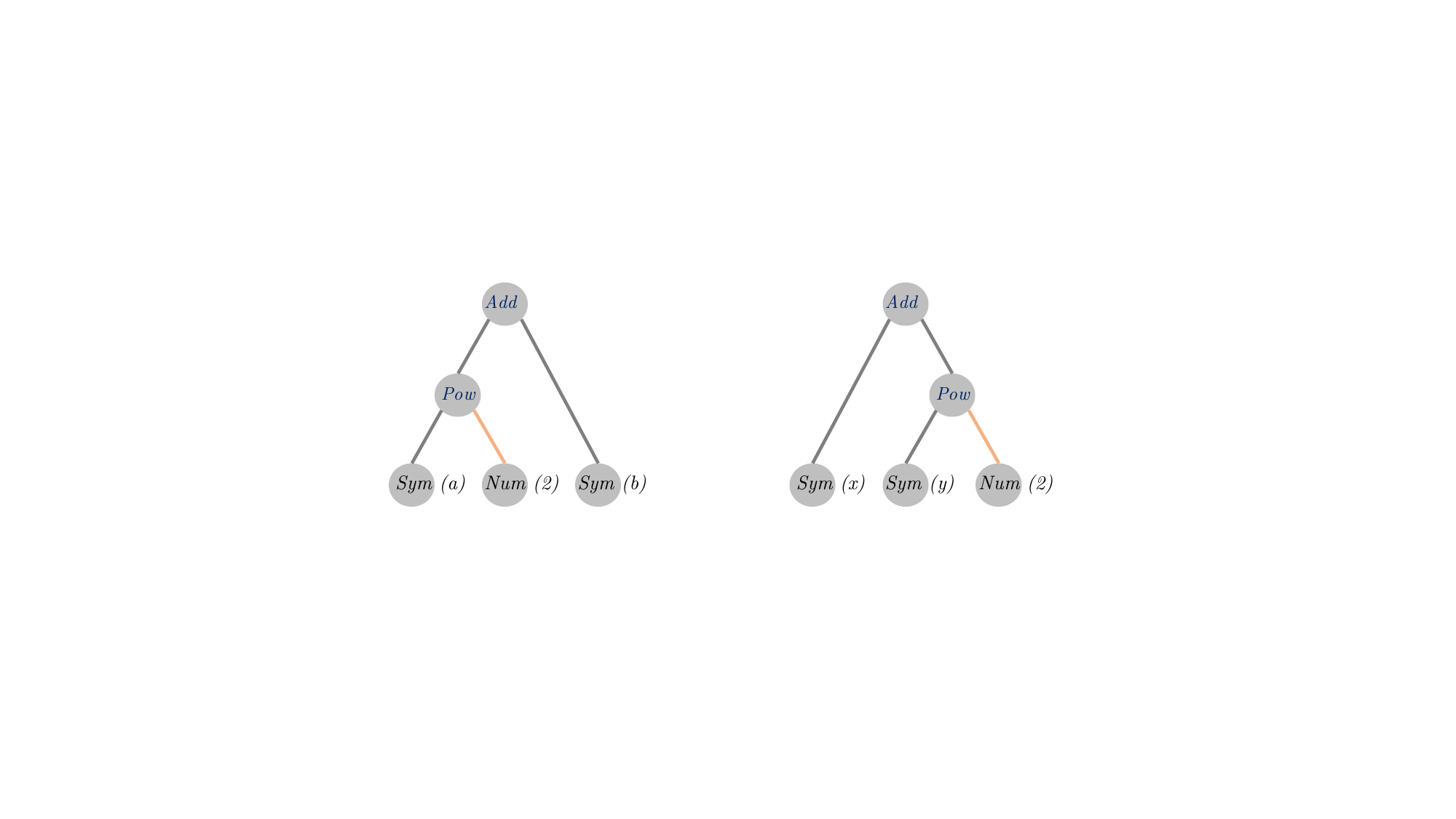}
\caption{Isomorphic expression tree for $a^2 + b$ and $x + y^2$.}
\label{fig:xy_vs_ab}
\end{figure}

\definecolor{MyGray}{gray}{.9}

\begin{table}[bt]
\begin{center}
\rowcolors{1}{white}{MyGray}
\begin{tabulary}{\textwidth}{R c p{9.2cm}}
Step 1  &:& rewrite expression in expanded form\\[1.35ex]
Step 2 &:& rewrite expression in prefix notation\\[1.35ex]
Step 3 &:& convert non-commutative operators\\[1.35ex]
Step 4 &:& build modified expression tree from prefix notation\\[1.35ex]
\hline
\end{tabulary}
\caption{Steps involved in obtaining an invariant representation for mathematical expressions.}
\label{tab:table}
\end{center}
\end{table}

\subsection{non-commutative binary operators}
Special care needs to be taken for binary operators whose operands do not commute. Consider $2-x$ and $x-2$. Encoded naively, their corresponding expression trees are going to be isomorphic, suggesting that the two expressions are the same, an obviously incorrect conclusion. To this end, subtraction is encoded as a combination of addition and multiplication by $-1$, both of which are commutative, i.e. $x-2$ becomes $x + (-1 \times 2)$.
Similarly, division is rewritten as multiplication with exponentiation: $\frac{x}{2} = x \times 2^{-1}$ whose expression tree will not be isomorphic to $\frac{2}{x} = 2 \times x^{-1}$. 

\begin{figure}[hbtp]
\includegraphics[trim=50 100 50 100, clip,scale=0.5]{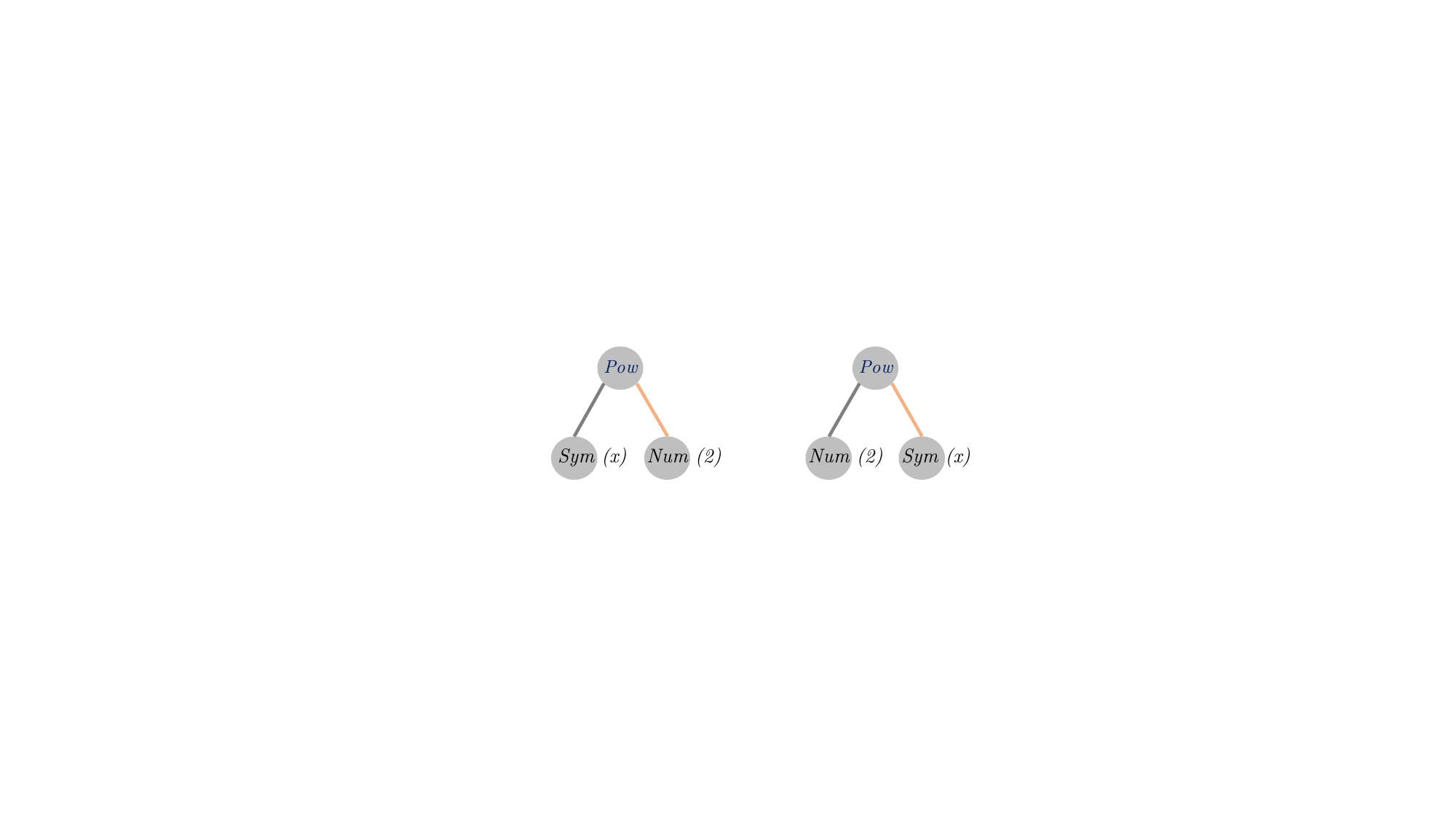}
\caption{The expressions trees for $2^x$ and $x^2$. Note that although the graphs are isomorphic, the expressions are not.}
\label{fig:noncommutative}
\end{figure}

Another such operation is ``Pow'', the operation that raises its base to its power. Figure \ref{fig:noncommutative} shows the corresponding graphs for $2^x$ and $x^2$. Once again, a naive encoding would result in isomorphic expression trees. Here, we differentiate the two edges extending from ``Pow'' to distinguish the base from the power. In figure \ref{fig:noncommutative} the power-edge is colored differently to reflect this differentiation. 

\subsection{Computational expense}
In practice, one may often need to compile a data base of all the mathematical expressions in a certain corpus, say, to support user queries. For instance, CK-12 may want to provide supporting material in response to users' queries. When those queries contain mathematical expressions, it would have to consult a data base compiled from all the expressions within the CK-12 corpus to retrieve the relevant documents. Considering that a \emph{match} in this scenario means finding a graph isomorphic to the query, the computational cost of searching the data base can be prohibitively high. In our experiments, we tackled this problem by assigning a \emph{key} to each graph, constructed by concatenating its nodes alphabetically sorted (for the two graphs in figure \ref{fig:xy_vs_ab} the key would be AddNumPowSymSym). These keys are then used as entry points into a hash table whose values are the list of all graphs with the same key (Note that since the key does not consider the structure of the graph, different graphs may have the same key). Effectively, during a search one only needs to examine the isomorphism of the graphs whose key matches the query: a significantly smaller number than the total number of graphs in the data base.

\section{Some Observations}

Here we summarize some of our observations from applying our encoding to the mathematical expressions within the CK-12 content, as well as their potential use cases.

\subsection{Building a Math-Based Concept Map}
\label{sec:conceptmap}
We used our algorithm to build a lookup table where the encoded expression serves as the key, and the stored values are the CK-12 concepts that the key appeared in. For instance, querying the table for $x^2+y^2=z^2$ you will get a list of concepts including
\begin{itemize}
\item Pythagorean Theorem and its Converse (Algebra),
\item Applications of Quadratic Equations (Algebra),
\item Circles Centered at the Origin (Calculus),
\item Graphical Methods of Vector Addition (Physics), and
\item Mechanical Advantage (Physics).
\end{itemize}
among others. One observation that immediately stands out is the natural linkages that the table forms among not just the various concepts within a discipline, but also among various disciplines (e.g. Algebra and Physics). This property can be used to build a mathematical concept map that relates various concepts to each other by virtue of their mathematical commonalities. Refer to table \ref{tab:conceptmap} for more examples of such a map.

\subsection{Uncovering Underlying Mathematical Constructs}
Looking at table \ref{tab:conceptmap} you will notice that the various concepts in the right column are linked via a specific expression (left). Put another way, in the lookup table discussed in section \ref{sec:conceptmap}, all the concepts that appear under a given key, are linked via that key, which is itself the encoded form of some mathematical expression. It follows that the ideas discussed in those concepts rely on that particular mathematical expression. In many cases that expression is only one among many, offering  only partial insight into the ideas discussed in that concept. However, there are also concepts that rely only on one or two main mathematical expressions which for the most part fully describe the mathematical aspects of the ideas discussed in those concepts. The interesting observation is that not only many Physics concepts are of this latter form, relying on one or two main mathematical expressions, but also that in many cases, conceptually unrelated physics concepts rely on the same mathematical expression. A simple example is the cluster of Physics concepts that are linked via the expression $\frac{\alpha}{\beta}$. This cluster includes concepts as diverse as Speed ($\frac{\text{distance}}{\text{time}}$), Density ($\frac{\text{mass}}{\text{volume}}$), Current ($\frac{\text{voltage}}{\text{resistance}}$), and Pressure ($\frac{\text{force}}{\text{area}}$).

What's more, these underlying mathematical constructs are not necessarily as simple as $\frac{\alpha}{\beta}$ and can take on more involved forms such as $k\frac{xy}{z^2}$ which is shared by both Coulomb's Law ($k\frac{q_1 q_2}{r^2}$) and Newton's formulation of Gravity ($G\frac{m_1 m_2}{d^2}$).

\subsection{As a means of Diagnosis of Knowledge Gaps}
A student's failure to fully grasp a new concept can be attributed to a number of factors. Over the years teachers come to develop intuitions about what the common culprits for a given concept are. For instance, in failing to understand the analysis of asymptotic behavior of rational expressions, teachers may have learned that the difficulty often arises from student's faulty memory of the mechanics of long division. In light of our discussion above on recovering the underlying mathematical constructs that figure prominently in certain concepts, we point out that our encoding system can be quite handy in quickly highlighting candidate mathematical concepts poor knowledge of which may be underpinning a student's struggle with a new concept. A student's struggle with the concept of Linear Regression may very well be due to conceptual misunderstandings which can hopefully be remedied by studying the material more carefully. However, it may also be the case that while the student understands the concept of linear regression well, she struggles with the mathematics of the equation of a line (see table \ref{tab:conceptmap} first row).

\definecolor{MyGray}{gray}{.9}

\begin{table}[bt]
\begin{center}
\rowcolors{1}{white}{MyGray}
\begin{tabulary}{\textwidth}{c c p{12cm}}

\textbf{Linked via} &:& \textbf{Linked Concepts}\\
\hline
$y = ax +b$  &:& Equations of Lines from Two Points (Algebra); Least-Squares Regression (Statistics); Instantaneous Rates of Change (Calculus); Specific Heat Calculations (Chemistery); Velocity and Acceleration (Physics); Motion (Physics)\\[1.35ex]

$\frac{a}{b}$&:& Slopes of Lines from Graphs (Algebra); Conditional Probability (Probability); Newton's Second Law (Physics); Position and Displacement (Physics); Velocity and Acceleration (Physics)\\[1.35ex]

$(\alpha+\beta)^2$ &:& Completing the Square (Algebra); Polynomial Expansion and Pascal's Triangle (Probability); Pythagorean Theorem and its Converse (Geometry); Implicit Differentiation (Calculus)\\[1.35ex]
\hline
\end{tabulary}
\caption{Examples of concepts linked via mathematical expressions.}
\label{tab:conceptmap}
\end{center}
\end{table}

\section{Discussion}

The basic implementation of this coding scheme is truly invariant to both variables and constants. As long as the structure of the expression is the same, it cares not what the specifics of the variables or constants are and treats them all the same. For instance, $(a+b)^7$ and $(u + v)^5$ are both coded the same way. While this may be desired for the purpose of measuring similarity, it can pose a problem when the specific variable or constant used bears a significance not shared by other choices for that variable or constant. Examples are variables of particular meaning in Physics (e.g. $R,C,V,G,\rho, \epsilon$ etc.), exponents of $2$ and $3$ which are generally treated separately and given special attention (e.g. many books cover $(a+b)^2$ and $(a+b)^3$ in isolation before generalizing to $(a+b)^n$), and constants such as $3.14$ and $9.8$. Therefore, it is important for the designer to consider carefully how much invariance is desired and include in the algorithm a list of special cases that require a different treatment.

\bibliographystyle{chicago}	
\bibliography{mybibs}	

\begin{thebibliography}{}

\bibitem[\protect\citeauthoryear{Arvind, Das, and K{\"o}bler}{Arvind
  et~al.}{2008}]{arvind2008logspace}
Arvind, V., B.~Das, and J.~K{\"o}bler (2008).
\newblock A logspace algorithm for partial 2-tree canonization.
\newblock In {\em International Computer Science Symposium in Russia}, pp.\
  40--51. Springer.

\bibitem[\protect\citeauthoryear{Devlin, Chang, Lee, and Toutanova}{Devlin
  et~al.}{2018}]{devlin2018bert}
Devlin, J., M.-W. Chang, K.~Lee, and K.~Toutanova (2018).
\newblock Bert: Pre-training of deep bidirectional transformers for language
  understanding.
\newblock {\em arXiv preprint arXiv:1810.04805\/}.

\bibitem[\protect\citeauthoryear{Weininger, Weininger, and Weininger}{Weininger
  et~al.}{1989}]{weininger1989smiles}
Weininger, D., A.~Weininger, and J.~L. Weininger (1989).
\newblock Smiles. 2. algorithm for generation of unique smiles notation.
\newblock {\em Journal of chemical information and computer sciences\/}~{\em
  29\/}(2), 97--101.

\end{thebibliography}

\end{document}